\documentclass{kybernetika}

\usepackage{makeidx}         
\usepackage{graphicx}        
\usepackage{multicol}        
\usepackage[bottom]{footmisc}
\usepackage{listings,color,algorithmic,algorithm}

\makeindex             

\newtheorem{theorem}{Theorem}[section]

\newtheorem{proposition}[theorem]{Proposition}

\newtheorem{definition}[theorem]{Definition}

\begin{document}
\pagestyle{myheadings}

\title{Towards an extension of the 2-tuple linguistic model to deal with unbalanced linguistic term sets}

\author{Mohammed-Amine ABCHIR and Isis TRUCK}

\contact{Mohammed-Amine}{Abchir}{CHArt -- EA4004, Universit\'e Paris
8, 2 rue de la Libert\'e, F-93526, Saint-Denis; Deveryware, 43 rue
Taitbout, F-75009 Paris. France.}{maa@ai.univ-paris8.fr}
\contact{Isis}{Truck}{CHArt -- EA4004, Universit\'e Paris 8, 2 rue de
la Libert\'e, F-93526, Saint-Denis.
France.}{isis.truck@univ-paris8.fr}

\markboth{Mohammed-Amine ABCHIR and Isis TRUCK} {Towards an extension of the 2-tuple model}

%
%
\maketitle

\begin{abstract}
In the domain of \emph{Computing with words} (CW), fuzzy linguistic approaches are known to be relevant in many decision-making
problems. Indeed, they allow us to model the human reasoning in replacing words, assessments, preferences, choices, wishes$\ldots$
by \emph{ad hoc} variables, such as fuzzy sets or more sophisticated variables.

This paper focuses on a particular model: Herrera \& Mart\'{i}nez' 2-tuple linguistic model and their approach to deal with unbalanced
linguistic term sets. It is interesting since the computations are accomplished without loss of information while the results
of the decision-making processes always refer to the initial linguistic term set. They propose a fuzzy partition which distributes
data on the axis by using linguistic hierarchies to manage the non-uniformity.
However, the required input (especially the density around the terms) taken by their fuzzy partition algorithm may be considered as too much demanding in a real-world
application, since density is not always easy to determine. Moreover, in some limit cases (especially when two terms are very closed semantically to each other), the partition doesn't comply with the data themselves, it isn't close to the reality.
Therefore we propose to modify the required input, in order to offer a simpler and more faithful partition.
We have added an extension to the package jFuzzyLogic and to the corresponding script language FCL.
This extension supports both 2-tuple models: Herrera \& Mart\'{i}nez' and ours.
In addition to the partition algorithm, we present two aggregation algorithms: the arithmetic means and the addition.
We also discuss these kinds of 2-tuple models.
\end{abstract}

\section{Introduction}
\label{intro}

Decision making is one of the most central human activities. The need of choosing between solutions in our complex world implies setting priorities on them considering multiple criteria such as benefits, risk, feasibility\dots~The interest shown by scientists to Multi Criteria Decision Making (MCDM) problems, as the survey of Bana e Costa shows~\cite{COSTA90}, has led to the development of many MCDM approaches such as the Utility Theory, Bayesian Theory, Outranking Methods and the Analytic Hierarchy Process (AHP). But the main lack of these approaches is that they represent the preferences of the decision maker about a real-world problem in a crisp mathematical model. As we are dealing with human reasoning and preference modeling, qualitative data and linguistic variables may be more suitable to represent linguistic preferences and their underlying aspects~\cite{Cha10}. 
Mart\'inez \emph{et al.} have presented in \cite{Mar10} a wide list of applications to show the usability and the advantages that the linguistic information
 (using various linguistic computational models) produce in decision making.
The preference extraction can be done thanks to elicitation strategies performed through User Interfaces (UIs)~\cite{Boo89} and Natural Language Processing (NLP)~\cite{Amb97} in a stimulus-response application for instance.

In the literature, many approaches allow to model the linguistic preferences and the interpretation made of it such as the classical fuzzy approach from Zadeh~\cite{Zad75}.
Zadeh has introduced the notions of linguistic variable and \emph{granule}~\cite{Zad97} as basic concepts that underlie human cognition. In~\cite{HACH09}, the authors review the computing with words in Decision Making and explain
that a granule ``which is the denotation of a word (\dots) is viewed as a fuzzy constraint on a variable''.

Among the existing models, there is one that permits to deal with granularity and 
with linguistic assessments in a fuzzy way with a simple and regular representation: the fuzzy linguistic 2-tuples introduced by Herrera and Mart\'inez~\cite{Her00a}.
Moreover, this model enables the representation of unbalanced linguistic data (\emph{i.e.} the fuzzy sets representing the terms are not symetrically and uniformly distributed on their axis).
However, in practice, the resulting fuzzy sets 
do not match exactly with human preferences.
Now we know how crucial the selection of the membership functions is to determine the
validity of a CW approach~\cite{Mar10}.
That is why an intermediate representation model is needed when we are dealing with data that are
``very unbalanced'' on the axis. 

The aim of this paper is to introduce another kind of fuzzy partition for unbalanced term sets, based on the fuzzy linguistic 2-tuple
model. Using the levels of linguistic hierarchies, a new algorithm is presented to improve the matching of the fuzzy partitioning.

This paper is structured as follows. First, we shortly recall the fuzzy linguistic approach and the 2-tuple fuzzy linguistic representation model
by Herrera \& Mart\'inez. In Section \ref{sec:ouralgo} we introduce a variant version of fuzzy linguistic 2-tuples and the corresponding partitioning algorithm before presenting aggregation operators (Section \ref{sec:aggreg}). Then in Section \ref{sec:discussion} another extension of the model and a prospective application of this new kind of 2-tuples are
discussed. We finally conclude with some remarks.

\section{The 2-tuple fuzzy linguistic representation model}
\label{sec:stateart}

In this section we remind readers of the fuzzy linguistic approach, the 2-tuple fuzzy linguistic representation model and some related works. We also review some studies on the use of natural language processing in human computer interfaces.

\subsection{2-tuples linguistic model and fuzzy partition}

Among the various fuzzy linguistic representation models, the approach that fits our needs the most is the representation that has been introduced by Herrera and Mart{\'{\i}}nez in~\cite{Her00a}.
This model represents linguistic information by means of a pair $(s, \alpha)$, where $s$ is a label representing the linguistic term and $\alpha$ is the value of the symbolic translation.
The membership function of $s$ is a triangular fuzzy set.

Let us note that in this paper we call a linguistic \emph{term} a word (\emph{e.g.} tall) and
a \emph{label} a symbol on the axis (\emph{i.e.} an $s$).

The computational model developed for this representation one includes comparison, negation and aggregation operators.
By default, all triangular fuzzy sets are uniformly distributed on the axis, but the targeted aspects are not usually uniform. In such cases, the representation should be enhanced with tools such as \emph{unbalanced} linguistic term sets which are not uniformly distributed on the axis~\cite{Herrera08afuzzy}.
To support the non-uniformity of the terms (we recall that the term set shall be unbalanced), the authors have chosen to change
the scale granularity, instead of modifying the shape of the fuzzy sets. The key element that manages multigranular linguistic information
is the \emph{level} of a \emph{linguistic hierarchy}, composed of an odd number of triangular fuzzy sets of the same shape, equally distributed on the axis, as a fuzzy partition in Ruspini's sense~\cite{Rus69}.

\begin{figure}[!h]
  \begin{center}
    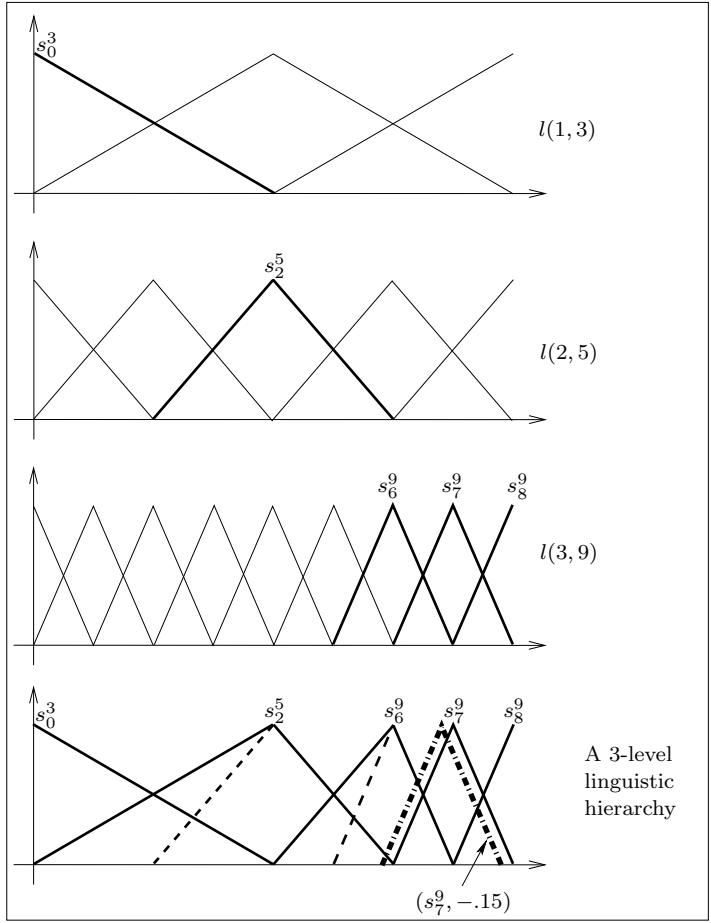
    \caption{\label{fig:2tuples}Unbalanced linguistic term sets: example of a 3 level-partition}
  \end{center}
\end{figure} 
A linguistic hierarchy $(LH)$ is composed of several label sets of different levels (\emph{i.e.}, with different granularities).
Each level of the hierarchy is denoted $l(t,n(t))$ where $t$ is the level number and $n(t)$ the number of labels (see Figure~\ref{fig:2tuples}).
Thus, a linguistic label set $S^{n(t)}$ belonging to a level $t$ of a linguistic hierarchy $LH$ can be denoted 
$S^{n(t)} = \{ s_0^{n(t)},\dots,s_{n(t)-1}^{n(t)} \}$.
In Figure~\ref{fig:2tuples}, it should be noted that $s_5^2$ (bottom, plain and dotted line) is a \emph{bridge unbalanced label} because
it is not symmetric. Actually each label has two sides: the upside (left side) that is denoted $\overline{s_i}$ and the downside (right side)
that is denoted $\underline{s_i}$. Between two levels there are \emph{jumps} so we have to bridge the unbalanced term to obtain
a fuzzy partition. Both sides of a bridge unbalanced label belong to two different levels of hierarchy.

\vspace{.5cm}
Linguistic hierarchies are unions of levels and assume the following properties~\cite{HER01}:

\begin{itemize}
\item levels are ordered according to their granularity;
\item the linguistic label sets have an odd number $n(t)$;
\item the membership functions of the labels are all triangular;
\item labels are uniformly and symmetrically distributed on $[0,1]$;
\item the first level is $l(1,3)$, the second is $l(2,5)$, the third is $l(3,9)$, etc.
\end{itemize}

Using the hierarchies, Herrera and Mart\'inez have developed an algorithm that permits to partition data in a convenient way.

This algorithm needs two inputs: the linguistic term set $\mathcal{S}$\footnote{When talking about linguistic terms, $\mathcal{S}$ (calligraphic font) is used, otherwise $S$ (normal font) is used.} (composed by the medium term denoted $\mathcal{S}_C$, the set of terms on its left denoted $\mathcal{S}_L$ and the set of terms on its right denoted $\mathcal{S}_R$) and the density of term distribution on each side. The density can be \emph{middle} or \emph{extreme} according to the user's choice. For example the description of $\mathcal{S} = \{A,B,C,D,E,F,G,H,I\}$ is $\{(2,extreme),1,(6,extreme)\}$ with $\mathcal{S}_L = \{A,B\}$, $\mathcal{S}_C = \{C\}$ and $\mathcal{S}_R = \{D,E,F,G,H,I\}$.

\subsection{Drawbacks of the 2-tuple linguistic model fuzzy partition in our context}

First, the main problem of this algorithm is the density. Since the user is not an expert, how could he manage to give the density? First,
he should be able to understand notions of granularity and unbalanced scales.

Second, it is compulsory to have an odd number of terms (\emph{cf.} $n(t)$)
in order to define a middle term (\emph{cf.} $\mathcal{S}_C$). But it may happen that the parity shall not be fulfilled. For example, when talking about a GPS
battery we can consider four levels: full, medium, low and empty.

Last, the final result may be quite different from what was initially expected because only a ``small unbalance'' is allowed. It means that even
if the \emph{extreme} density is chosen, it doesn't guarantee the obtention of a very thin granularity. Only two levels of density are allowed
(\emph{middle} or \emph{extreme}) which can be a problem when considering distances such as: arrived, very closed, closed, out of reach. ``Out of reach'' needs a level of granularity quite different from the level for terms ``arrived'', ``very closed'' and ``closed''.

As the fuzzy partition obtained by this approach does not always fit with the reality, we proposed in~\cite{MAA11} a draft of approach to overcome this problem.
This is further described in~\cite{MAAIT11} where we mainly focus on the industrial context (geolocation) and the underlying problems addressed by our
specific constraints.

The implementations and tests made for this work are based on the jFuzzyLogic library. It is the most used fuzzy logic package by Java developers. It implements Fuzzy Control Language (FCL) specification (IEC 61131-7) and is available under the Lesser GNU Public Licence (LGPL).

Even if it is not the main point of this paper, one part of our work is to provide an interactive tool in the form of a natural language dialogue interface. This dialogue, through an elicitation strategy, helps to extract the human preferences. We use NLP
techniques to represent the grammatical, syntactical and semantic relations between the words used during the interaction part.
Moreover, to be able to interpret these words, the NLP is associated to fuzzy linguistic techniques. Thus, fuzzy semantics are associated to each word which is supported by the interactive tool (especially adjectives such as ``long'', ``short'', ``low'', ``high'', etc.) and can be used at the interpretation time.
This NLP-Fuzzy Linguistic association also enables to assign different semantics to the same word depending on the user's criteria (business domain, context, etc.). It allows then to unify the words used in the dialogue interface for different use cases by only switching between their different semantics.

Another interesting aspect of this NLP-fuzzy linguistic association lies in the possibility of an automatic semantic generation in a sort of autocompletion mode.

For example, in a geolocation application, if the question is ``\emph{When do you want to be notified?}'', a user's answer can be ``\emph{I want to be notified when the GPS battery level is \textbf{low}}''. Here the user says \emph{low}, so we propose a semantic distribution of the labels of the term set according to the number of the synonyms of this term. Indeed, the semantic relations between words introduced by NLP (synonyms, homonyms, opposites, etc.) can be used to highlight words associated with the term \emph{low} semantically and then to construct a linguistic label set around it.
The more relevant words found for a term, the higher the density of labels is around it. In comparison with the 2-tuple fuzzy linguistic model introduced by Herrera \& al., this amounts to deduce the \emph{density} (in Herrera \& Mart\'inez' sense) according to the number of
synonyms of a term.
In practice, thanks to a synonym dictionary it is possible to compute a semantic distance between
each term given by the geolocation expert. If two terms are considered as synonymous
they will share the same $LH$. Moreover, a word with few (or no) synonyms will be represented
in a coarse-grained hierarchy while a word with many synonyms will be represented in a fine-grained
hierarchy.

We can see here how much the unbalanced linguistic label sets can be relevant in many situations.
To couple NLP techniques and fuzzy linguistic models seems very promising.


\section{Towards another kind of 2-tuples linguistic model}
\label{sec:ouralgo}

Starting from a running example, we now present our proposal that aims at avoiding the drawbacks mentioned above.

\subsection{Running example}
   
\begin{figure}[t]
  \begin{center}
    \includegraphics[width=12cm]{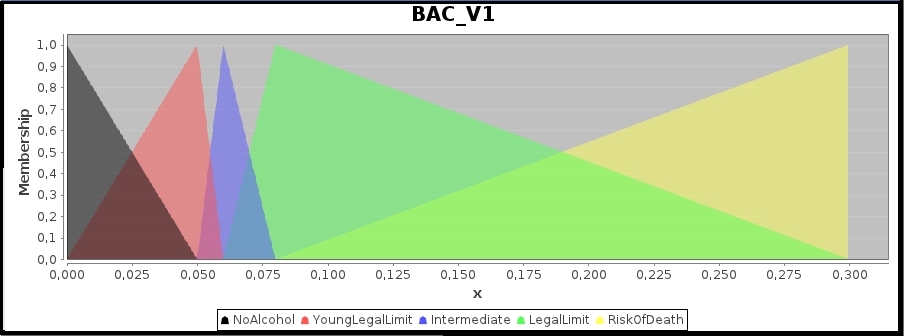}
  \end{center}
  \caption{The ideal fuzzy partition for the BAC example.}
  \label{fig:idealPart}
\end{figure}

Herrera \& Mart\'{i}nez' methodology needs a term set $\mathcal{S}$ and an associated description with two densities. For instance, when
considering the blood alcohol concentration (BAC in percentage) in the USA, we can focus on five main values:
$0\%$ means no alcohol, $.05\%$ is the legal limit for drivers under 21, $.065\%$ is an intermediate value (illegal
for young drivers but legal for the others), $.08\%$ is the legal limit for drivers older than 21 and $.3\%$ is
considered as the BAC level where risk of death is possible. In particular, the ideal partition should comply with the data and with the gap
between values (see Figure~\ref{fig:idealPart} that simply proposes triangular fuzzy sets without any real semantics, obtained
directly from the input values). But this prevents us from using the advantages of Herrera \& Mart\'{i}nez' method, that are
mainly to keep the original semantics of the terms, \emph{i.e.} to keep the same terms from the
original linguistic term set. The question is how to express linguistically the results of the computations if the partition doesn't fulfill
``good'' properties such as those from the 2-tuple linguistic model?

\subsection{Extension of jFuzzyLogic and preliminary definitions}
\label{ssec:model}

With Herrera \& Mart\'{i}nez' method, we have\\
$\mathcal{S} = \{\textit{NoAlcohol},$ $\textit{YoungLegalLimit},$ $\textit{Intermediate},$ $\textit{LegalLimit},$ $\textit{RiskOfDeath}\}$ and its description
is $\{(3,extreme),1,(1,extreme)\}$
with $\mathcal{S}_L = \{\textit{NoAlcohol}$, $\textit{YoungLegalLimit}$, $\textit{Intermediate}\}$, $\mathcal{S}_C = \{\textit{LegalLimit}\}$ and
$\mathcal{S}_R = \{\textit{RiskOfDeath}\}$.

jFuzzyLogic extension  (we have added the management of Herrera \& Mart\'inez'
2-tuple linguistic model) helps modeling this information and we obtain the following FCL script:

\noindent
\verb?VAR_INPUT?\\
\verb?       BloodAlcoholConcentration : LING;?\\
\verb?END_VAR?\\

\noindent
\verb?FUZZIFY BloodAlcoholConcentration?\\
\verb?       TERM S := ling NoAlcohol YoungLegalLimit?\\
\verb?             Intermediate | LegalLimit | RiskOfDeath,?\\
\verb?             extreme extreme;?\\
\verb?END_FUZZIFY?\\

\begin{figure}[h]
  \begin{center}
    \includegraphics[width=12cm]{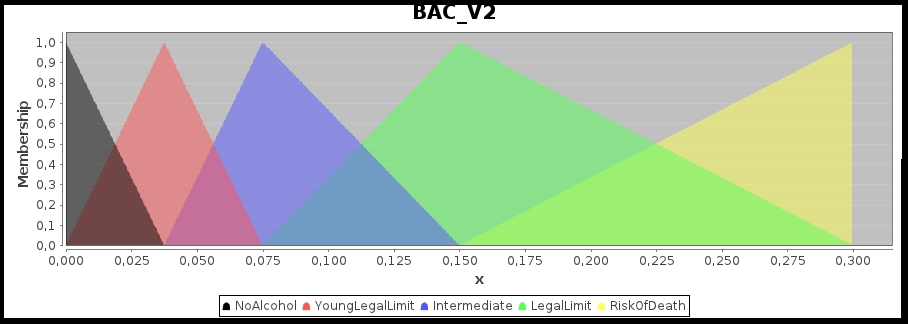}
  \end{center}
  \caption{Fuzzy partition generated by Herrera \& Mart\'{i}nez' approach.}
  \label{fig:partLuis}
\end{figure}

The resulting fuzzy partition is quite different from what was initially expected (see Figure~\ref{fig:partLuis} compared to
Figure~\ref{fig:idealPart} where we notice that the label unbalance is not really respected).
We recall that each label $s_i$ has two sides. For instance, the label $s_i$ associated to \textit{NoAlcohol} has a downside
and no upside
while the term $s_j$ associated to \textit{RiskOfDeath} has an upside and no downside.

Two problems appear: the use of densities is not always obvious for final users, and the gaps between values (especially between
\textit{LegalLimit} and \textit{RiskOfDeath}) are not respected.

To avoid the use of the densities that can be hard to obtain from the user (\emph{e.g.}, see the specific geolocation
industrial context explained in~\cite{MAAIT11}),
we have evoked in~\cite{MAA11} a tentative approach which offers a simpler way to retrieve unbalanced linguistic terms. The aim was to accept
any kind of description of the terms coming from the user. That is why we propose an extension of jFuzzyLogic to handle linguistic 2-tuples in addition to an
enrichment of the FCL language specification. Consequently, we suggest another way to define a \texttt{TERM} with a new type of variable called \texttt{LING}
(see the example below).
\\

\noindent
\verb?VAR_INPUT?\\
\verb?       BloodAlcoholConcentration : LING;?\\
\verb?END_VAR?\\

\noindent
\verb?FUZZIFY BloodAlcoholConcentration?\\
\verb?       TERM S := ling (NoAlcohol,0.0) (YoungLegalLimit,0.05)?\\
\verb?             (Intermediate,0.065) (LegalLimit,0.08) (RiskOfDeath,0.3);?\\
\verb?END_FUZZIFY?\\

It should be noted that the linguistic values are composed by a pair $(\mathsf{s,v})$ where $\mathsf{s}$ is a linguistic term (\emph{e.g.}, \textit{LegalLimit})
and $\mathsf{v}$ is a number giving the position of $\mathsf{s}$ on the axis (\emph{e.g.}, $0.08$).
Thus several definitions can now be given.

\begin{definition} \label{def:SRonde}
  Let $\mathcal{S}$ be an unbalanced ordered linguistic term set and $U$ be the numerical universe where the terms are projected.
  Each linguistic value is defined by a unique pair $(\mathsf{s,v}) \in \mathcal{S} \times U$. The numerical distance between $\mathsf{s}_i$
  and $\mathsf{s}_{i+1}$ is denoted by $d_i$ with $d_i=\mathsf{v}_{i+1}-\mathsf{v}_i$.
\end{definition}

\begin{definition}
  Let $S=\{s_0,\ldots,s_p\}$ be an unbalanced linguistic label set and $(s_i,\alpha)$ be
  a linguistic 2-tuple.  To support the unbalance, $S$ is extended to several balanced linguistic label sets, each one
  denoted $S^{n(t)}=\{s_0^{n(t)},\ldots,s_{n(t)-1}^{n(t)}\}$
  (obtained from the algorithm of~\cite{HER01}) defined in the level $t$
  of a linguistic hierarchy $LH$ 
  with
  $n(t)$ labels. There is a unique way to go from $\mathcal{S}$ (Definition \ref{def:SRonde}) to $S$, according to Algorithm~\ref{algo_pa}.
\end{definition}

\begin{definition}
  Let $l(t,n(t))$ be a level from a linguistic hierarchy. The \emph{grain} $g$ of $l(t,n(t))$ is defined as the distance
  between two 2-tuples $(s_i^{n(t)},\alpha)$.
\end{definition}

\begin{proposition}
The grain $g$ of a level $l(t,n(t))$ is obtained as: $g_{l(t,n(t))}=1/(n(t)-1)$.
\end{proposition}

\begin{Proof}
  $g$ is defined as the distance between $(s_i^{n(t)},\alpha)$ and $(s_{i+1}^{n(t)},\alpha)$,
 \emph{i.e.}, between two kernels of the associated triangular fuzzy sets because $\alpha$ equals $0$.
Since the hierarchy is normalized on $[0,1]$, this distance is easy to compute using
$\Delta^{-1}$ operator from \cite{HER01} where $\Delta^{-1}(s_i^{n(t)},\alpha)=\frac{i}{n(t)-1}+\alpha=\frac{i}{n(t)-1}$.
 As a result, $g_{l(t,n(t))}=\frac{(i+1)}{n(t)-1}-\frac{i}{n(t)-1} = 1/(n(t)-1)$.
\end{Proof}
 
For instance, the grain of the second level is $g_{l(2,5)}=.25$.

\begin{proposition}\label{2*grain}
The grain $g$ of a level $l(t-1,n(t-1))$ is twice the grain of the level $l(t,n(t)$: $g_{l(t-1,n(t-1))} = 2g_{l(t,n(t))}$
\end{proposition}

\begin{Proof}
  This comes from the following property of the linguistic hierarchies.
  Let $l(t,n(t))$ be a level. Its successor is defined as: $l(t+1,2n(t)-1)$ (see~\cite{Herrera08afuzzy}).
\end{Proof}

\subsection{A new partitioning}
\label{ssec:newpart}




The aim of the partitioning is to assign a label $s_i^{n(t)}$ (indeed one or two) to each term $\mathsf{s}_k$. The selection of
$s_i^{n(t)}$ depends on both the distance $d_k$ and the numerical value $\mathsf{v}_k$.
We look for the nearest level --- they are all known in advance, see Table 1 in~\cite{Herrera08afuzzy} ---
\emph{i.e.}, for the level with the closest grain from $d_k$.
Then the right $s_i^{n(t)}$ is chosen to match $\mathsf{v}_k$ with the best accuracy. $i$ has to minimize the quantity
$\min_i |\Delta^{-1}(s_i^{n(t_k)},0)-\mathsf{v}_k|$.

By default, the linguistic hierarchies are distributed on $[0,1]$, so a scaling is needed in order
that they match the universe $U$.

The detail of these different steps is given in Algorithm~\ref{algo_pa}.
We notice that there is \emph{no condition} on the \emph{parity} of the number of terms.
Besides, the function returns a set of bridge unbalanced linguistic 2-tuples with a level of granularity that may not be the same
for the upside than for the downside.

\begin{algorithm}[t]
\caption{Partitioning algorithm}
\label{algo_pa}
\begin{algorithmic}[1]
\REQUIRE $\langle(\mathsf{s}_0,\mathsf{v}_0), \ldots, (\mathsf{s}_{p-1},\mathsf{v}_{p-1})\rangle$ are $p$ pairs of $\mathcal{S} \times U$;\\
         $t, t_0, \ldots, t_{p-1}$ are levels of hierarchies
\STATE scale the linguistic hierarchies on $[0,\mathsf{v}_{\textit{max}}]$, with $\mathsf{v}_{\textit{max}}$ the maximum $\mathsf{v}$ value
\STATE precompute $\eta$ levels and their grain $g$ ($\eta \geq 6$)

\FOR {$k=0$ to $p-1$}
   \STATE $d_k \gets \mathsf{v}_{k+1} - \mathsf{v}_{k}$
   \FOR {$t=\eta$ to $1$}
      \IF {$g_{l(t,n(t))} \leq d_k$} 
        \STATE $t_k \gets t$
      \ENDIF
   \ENDFOR
   \STATE $\textit{tmp}=\mathsf{v}_{\textit{max}}$
   \FOR {$i=0$ to $n(t_k)-1$}
      \IF {$\textit{tmp} > |\Delta^{-1}(s_i^{n(t_k)},0)-\mathsf{v}_k|$}
         \STATE $\textit{tmp} = |\Delta^{-1}(s_i^{n(t_k)},0)-\mathsf{v}_k|$
         \STATE $j \gets i$
      \ENDIF
   \ENDFOR
   \STATE $\underline{s_k^{n(t_k)}} \gets \underline{s_{j}^{n(t_k)}}$ ;
          $\overline{s_{k+1}^{n(t_k)}} \leftarrow \overline{s_{j+1}^{n(t_k)}}$
   \STATE depending on the level, $\underline{\alpha_k}=\mathsf{v}_k-\Delta^{-1}(s_j^{n(t_k)},0)$ or\\
 ~~~~~~~~~~~~~~~~~~~~~~~~~~~~~~~
          $\overline{\alpha_{k+1}}=\mathsf{v}_{k+1}+\Delta^{-1}(s_{j+1}^{n(t_{k})},0)$
\ENDFOR
\RETURN the set $\{ (\underline{s_{0}^{n(t_0)}}, \underline{\alpha_0}), (\overline{s_{1}^{n(t_0)}}, \overline{\alpha_1}),
 (\underline{s_{1}^{n(t_1)}}, \underline{\alpha_1}),\ldots,$\\
 ~~~~~~~~~~~~~~~~~~~~~~
$(\underline{s_{p-2}^{n(t_{p-2})}}, \underline{\alpha_{p-2}}),
 (\overline{s_{p-1}^{n(t_{p-2})}}, \overline{\alpha_{p-1}})\}$
\end{algorithmic}
\end{algorithm}

Herrera \& Mart\'{i}nez' partitioning does not follow exactly the user wishes because it transforms them into a model with
many properties, such as Ruspini conditions~\cite{Rus69}. As for us, we try to match the wishes as best as possible by
adding lateral translations $\alpha$ to the labels $s_i^{n(t)}$.  From this, it results a possible non-fulfillment of the previous properties.
For instance, what we obtain is not a fuzzy partition.
But we assume to do without these conditions since the goal is to totally cover the universe.
This is guaranteed by the \emph{minimal covering property}.

\label{appartMinimale}

\begin{proposition}
The 2-tuples $(s_i^{n(t)},\alpha)$ (from several levels $l(t,n(t))$) obtained from our partitioning algorithm are triangular fuzzy sets that cover
the entire universe $U$.
\end{proposition}

Actually, the distance between any pair $\langle(\underline{s_k^{n(t)}},\underline{\alpha_k}),
(\overline{s_{k+1}^{n(t)}},\overline{\alpha_{k+1}})\rangle$
is always strictly greater than twice the grain of the corresponding level.

\begin{Proof}
By definition and construction, $d_k$ is used to choose the convenient level $t$ for this pair.
We recall that when $t$ decreases, $g_{l(t,n(t))}$ increases.
As a result, we have:
\begin{equation}
  \label{eq:dists}
  g_{l(t,n(t))} \leq d_k < g_{l(t-1,n(t-1))}
\end{equation}
After having applied the steps of the assignation process we obtain two linguistic 2-tuples $(\underline{s_k^{n(t)}}, \underline{\alpha_k})$ and $(\overline{s_{k+1}^{n(t)}}, \overline{\alpha_{k+1}})$ representing the downside and upside of labels $s_k^{n(t)}$ and $s_{k+1}^{n(t)}$ respectively. 

Thanks to the symbolic translations $\alpha$, the distance between the kernel of these two 2-tuples is $d_k$.
Then, according to Proposition~\ref{2*grain} and to Equation~\ref{eq:dists} we conclude that:
\begin{equation}
  \label{eq:lhProperty}
  d_k < 2g_{l(t,n(t))}
\end{equation}
which means that, for each value in $U$, this fuzzy partition has a minimum membership value $\varepsilon$ strictly greater than 0. 

Considering $\mu_{s_i^{n(t)}}$ the membership function associated with a label $s_i^{n(t)}$, this property is denoted: 
\begin{equation}
  \label{eq:coverage}
  \forall u \in U, \ \ \ \ \  \mu_{s_0^{n(t_0)}}(u) \vee \dots \vee \mu_{s_i^{n(t_i)}}(u) \vee \dots \vee \mu_{s_{p-1}^{n(t_{p-1})}}(u) \geq \varepsilon > 0
\end{equation}

\end{Proof}

\vspace{0.3cm}
   
To illustrate this work, we take the running example concerning the BAC.
The set of pairs $(\mathsf{s,v})$ is the following: $\{(\textit{NoAlcohol},.0)$, $(\textit{YoungLegalLimit},.05)$ $(\textit{Intermediate},.065)$
$(\textit{LegalLimit},.08)$ $(\textit{RiskOfDeath},.3)\}$.

It should be noted that our algorithm implies to add another level of hierarchy: $l(0,2)$.

We denote by $L$ and $R$ the upside and downside of labels respectively. Table 1 shows the results, with
$\alpha$ values not normalized. To normalize them, it is easy to see that they have to be multiplied
by $1/.3$ because $\mathsf{v}_\textit{max}=.3$.

\begin{table}[h!] 
\begin{center}
 \begin{tabular}{|l|l|l|}
  \hline\hline
  linguistic term & level & 2-tuple\\
  \hline\hline
  \textit{NoAlcohol\_R} & $l(3,9)$ & $(s_0^9,0)$\\
  \hline
  \textit{YoungLegalLimit\_L} & $l(3,9)$ &  $(s_1^9,.0125)$\\
  \hline
  \textit{YoungLegalLimit\_R} & $l(5,33)$ &  $(s_5^{33},.003)$\\
  \hline
  \textit{Intermediate\_L} & $l(5,33)$ &  $(s_6^{33},0)$\\
  \hline
  \textit{Intermediate\_R} & $l(4, 17)$ &  $(s_3^{17},0)$\\
  \hline
  \textit{LegalLimit\_L} & $l(4, 17)$ &  $(s_4^{17},.005)$\\
  \hline
  \textit{LegalLimit\_R} & $l(1, 3)$ &  $(s_1^{3},-.07)$\\
  \hline
  \textit{RiskOfDeath\_R} & $l(1, 3)$ &  $(s_1^3,0)$\\
  \hline\hline
 \end{tabular}
 \caption{The 2-tuple set for the BAC example.}
\end{center}
\end{table}

\begin{figure}[t]
  \begin{center}
    \includegraphics[width=12cm]{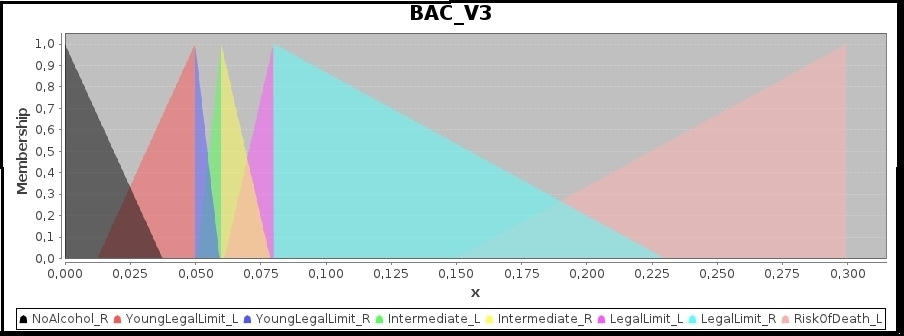}
  \end{center}
  \caption{Fuzzy partition generated by our algorithm for the BAC example.}
  \label{fig:ourPart}
\end{figure}

See Figure~\ref{fig:ourPart} for a graphical representation of the fuzzy partition.

\section{Aggregation with our 2-tuples}
\label{sec:aggreg}
\subsection{Arithmetic mean}
\label{ssec:aggreg}
   

As our representation model is based on the 2-tuple fuzzy linguistic one, we can use the aggregation operators (weighted average, arithmetic mean, etc.) of the unbalanced linguistic computational model introduced in~\cite{Herrera08afuzzy}. The functions $\Delta$, $\Delta^{-1}$, $\mathcal{LH}$ and $\mathcal{LH}^{-1}$ used in our aggregation are derived from the same functions in Herrera \& Mart\'inez' computational model.

In the aggregation process, linguistic terms $(\mathsf{s}_k, \mathsf{v}_k)$ belonging to a linguistic term set
$\mathcal{S}$ have to be dealt with. After the assignation process, these terms are associated to one or
two 2-tuples $(s_i^{n(t)}, \alpha_i)$ (remember the upside
and downside of a label) of a level from
a linguistic hierarchy $LH$.
We recall two definitions taken from~\cite{Herrera08afuzzy}.

\begin{definition}
$\mathcal{LH}^{-1}$ is the transformation function that associates with each linguistic 2-tuple expressed in $LH$
its respective unbalanced linguistic 2-tuple.
\end{definition}

\begin{definition}
Let $S= \{s_0,\ldots,s_g\}$ be a linguistic label set and $\beta \in [0,g]$ a value supporting the result of a symbolic aggregation
operation. Then the linguistic 2-tuple that expresses the equivalent information to $\beta$ is obtained with the function
$\Delta : [0,g] \longrightarrow S \times [-.5,.5)$, such that
\[\Delta(\beta)=\displaystyle\left\{\begin{tabular}{ll}
                                 $s_i$ & $i=\mathit{round}(\beta)$\\
                                 $\alpha=\beta - i$ & $\alpha \in [-.5, .5)$
                               \end{tabular}
                               \right.
\]
where $s_i$ has the closest index label to $\beta$ and $\alpha$ is the value of the symbolic translation.
\end{definition}

Thus the aggregation process (arithmetic mean) can be summarized by the three following steps:

\begin{enumerate}
\item Apply the aggregation operator to the $\mathsf{v}$ values of the linguistic terms. Let $\beta$ be the result of this aggregation.
\item Use the $\Delta$ function to obtain the $(s_q^r, \alpha_q)$ 2-tuple of $LH$ corresponding to $\beta$.
\item In order to express the resulting 2-tuple in the initial linguistic term set $\mathcal{S}$, we use the $\mathcal{LH}^{-1}$ function as defined in~\cite{Herrera08afuzzy} to obtain the linguistic pair $(\mathsf{s}_l, \mathsf{v}_l)$.
\end{enumerate}

\vspace{.3cm}

To illustrate the aggregation process, we suppose that we want to aggregate two terms (two pairs ($\mathsf{s},\mathsf{v}$))
of our running example concerning the BAC: (\emph{YoungLegalLimit}, .05) and (\emph{LegalLimit}, .08). In this example we use the arithmetic mean as aggregation operator.

Using our representation algorithm, the term (\emph{YoungLegalLimit}, .05) is associated to $(\underline{s_{1}^9}, .125)$ and $(\overline{s_5^{33}}, .003)$ and (\emph{LegalLimit}, .08) is associated to $(\underline{s_{4}^{17}}, .005)$ and $(\overline{s_1^{3}}, -.07)$.
First, we apply the arithmetic means to the $\mathsf{v}$ value of the two terms. As these values are in absolute scale, it simplifies the computations. The result of the aggregation is $\beta = .065$.

The second step is to represent the linguistic information of aggregation $\beta$ by a linguistic label expressed in $LH$. For the representation we choose the level associated to the two labels with the finest grain. In our example it is $l(5,33)$ (fifth level of $LH$ with $n(t)=33$). Then we apply the $\Delta$ function on $\beta$ to obtain the result: $\Delta(.065) = (s_7^{33}, -.001)$. 

Finally, in order to express the above result in the initial linguistic term set $\mathcal{S}$, we apply the $\mathcal{LH}^{-1}$ function. It associates to a linguistic 2-tuple in $LH$ its corresponding linguistic term in $\mathcal{S}$. Thus, we obtain the final result $\mathcal{LH}^{-1}((s_7^{33}, -.001)) =$ (\emph{YoungLegalLimit}, .005).

Given that countries have different rules concerning the BAC for drivers, the aggregation of such linguistic information can be relevant to calculate an average value of allowed and prohibited blood alcohol concentration levels for a set of countries (Europe, Africa, etc.).


\subsection{Addition}
\label{ssec:autreOper}

    
As we are using an absolute scale on the axis for our linguistic terms, the approach for other operators is the same as the one described above for the arithmetic means aggregation. We first apply the operator to the $\mathsf{v}$ values of the linguistic terms and then we use  the $\Delta$ and the $\mathcal{LH}^{-1}$ functions successively to express the result in the original term set.

If we consider for instance that, this time, we need to add the two following terms: (\emph{YoungLegalLimit}, .05) and (\emph{LegalLimit}, .08), we denote $(\emph{YoungLegalLimit}, .05) \oplus (\emph{LegalLimit}, .08)$ and
proceed as follows:

\begin{itemize}
\item We add the two $\mathsf{v}$ values $.05$ and $.08$ to obtain $\beta = .13$.
\item We then apply the $\Delta$ function to express $\beta$ in $LH$, $\Delta(0.13) = (s_{14}^{33}, -.001)$.
\item Finally, we apply the $\mathcal{LH}^{-1}$ function to obtain the result expressed in the initial linguistic term set $\mathcal{S}$ : $\mathcal{LH}^{-1}((s_{14}^{33}, -.001)) =$ (\emph{LegalLimit}, .05).
\end{itemize}

This $\oplus$ addition looks like a fuzzy addition operator (see \emph{e.g.} \cite{Her00a}) used as a basis for many aggregation
processes (combine experts' preferences, etc.). Actually, $\oplus$ operator can be seen as an extension (in the sense
of Zadeh's principle extension) of the addition for our 2-tuples.


The same approach can be applied to other operators. It will be further explored in our future works.

\section{Discussions}
\label{sec:discussion}

\subsection{Towards a fully linguistic model}
When dealing with linguistic tools, the aim is to avoid the user to supply precise numbers, since he's not always able to give them.
Thus, in the pair $(\mathsf{s},\mathsf{v})$ that describes the data, it may happen that the user doesn't know exactly the position $\mathsf{v}$.

\begin{figure}[!h]
  \begin{center}
    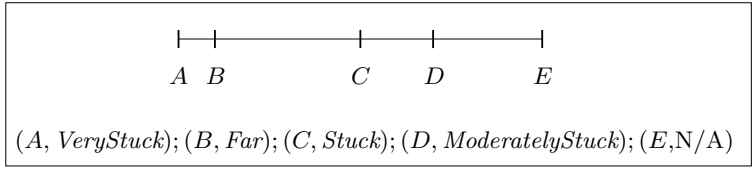
    \caption{\label{fig:StretchFactor}Example of the use of a stretch factor}
  \end{center}
\end{figure}

 For instance,
considering five grades $(A,B,C,D,E)$, the user knows that (i) $D$ and $E$ are fail grades, (ii) $A$ is the best one, (iii) $B$ is not far away,
(iv) $C$ is in the middle. If we replace $\mathsf{v}$ by a linguistic term, that is a \emph{stretch factor}, the five pairs in the previous example could be:
$(A,\textit{VeryStuck}); (B,\textit{Far}); (C,\textit{Stuck}); (D,\textit{ModeratelyStuck});$ $(E,$N/A$)$ (see Figure
\ref{fig:StretchFactor}). $(A,\textit{VeryStuck})$ means that $A$ is very stuck to its next label.
$(E,$N/A$)$ means that $E$ is the last label ($\mathsf{v}$ value is not applicable).

This improvement permits to ask the user for:

\begin{itemize}
  \item either the pairs $(\mathsf{s},\mathsf{v})$, with $\mathsf{v}$ a linguistic term (stretch factor);
  \item or only the labels $\mathsf{s}$ while placing them on a visual scale (\emph{i.e.}, the stretch factors are automatically computed to obtain the pairs
  $(\mathsf{s},\mathsf{v})$);
  \item or the pairs $(\mathsf{s},\mathsf{v})$, with $\mathsf{v}$ a numerical value, as proposed above.
\end{itemize}

It should be noted that the first case ensures to deal with fully linguistic pairs $(\mathsf{s},\mathsf{v})$. It should also be noted that our stretch factor looks like Herrera \& Mart\'{i}nez'
densities, but in our case, it permits to construct a more accurate representation of the terms.

\subsection{Towards a simplification of binary trees}

The linguistic 2-tuple model that uses the pair $(s_i^{n(t)},\alpha)$ and its corresponding level of linguistic hierarchy can be seen as another way to express
the various nodes of a tree. There is a parallel to draw between the node depth and the level of the linguistic hierarchy.
Indeed, let us consider a binary tree, to simplify. The root node belongs to the first level, that is $l(1,3)$ according to \cite{HER01}.
Then its children belong to the second one ($l(2,5)$), knowing that the next level is obtained from its predecessor:
$l(n+1,2n(t)-1)$. And so on, for each node, until there is no node left.
In the simple case of a binary tree (\emph{i.e.}, a node has two children or no child), it is easy to give the position --- the 2-tuple $(s_i^{n(t)},\alpha)$ ---
of each node: this position is unique, left child is on the left of its parent in the next level (resp. right for the right child).

\begin{figure}[!h]
  \begin{center}
    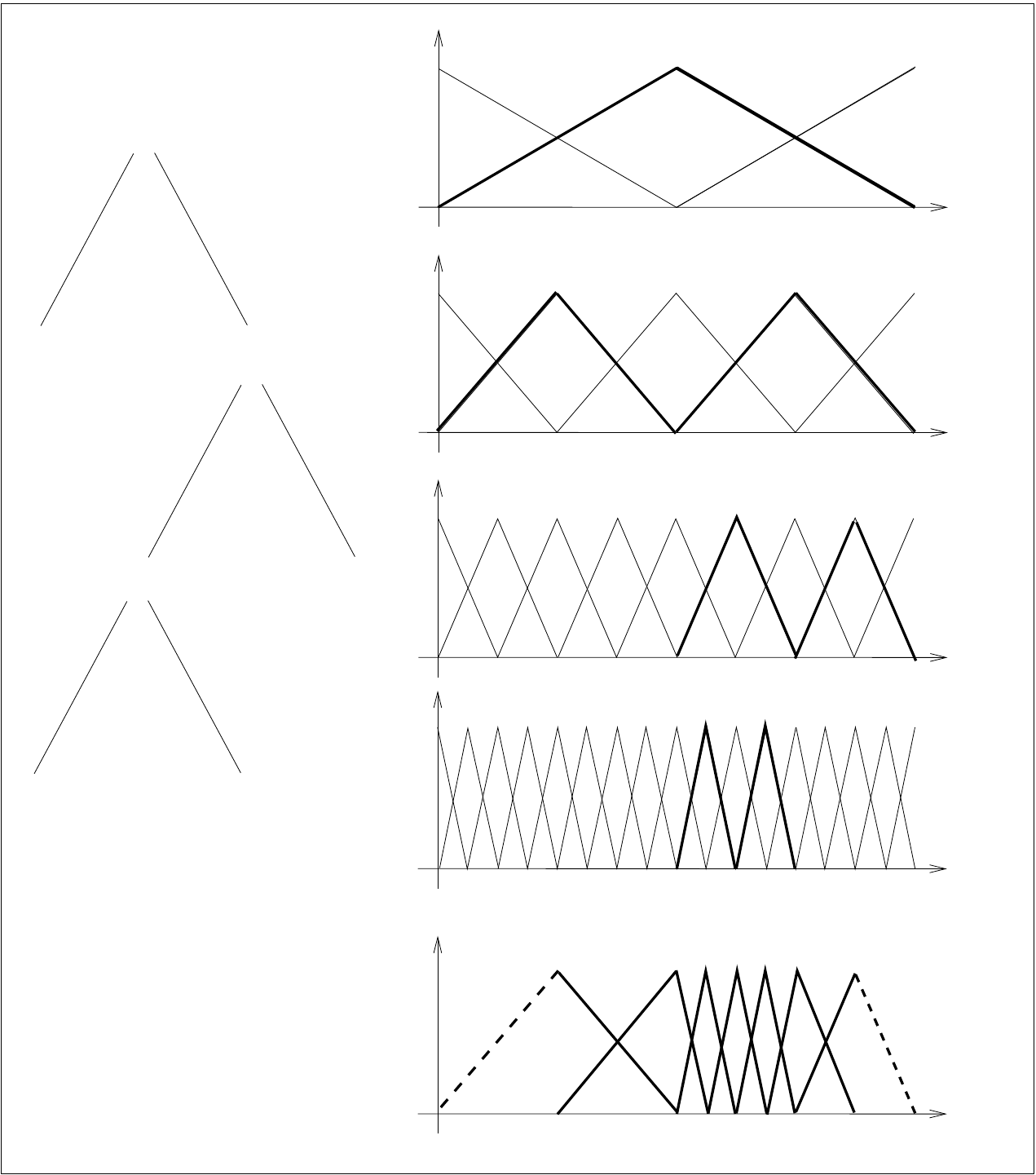
    \caption{\label{fig:simplif}Example of the simplification of a binary tree}
  \end{center}
\end{figure}

The algorithm that permits to simplify a binary tree in a linguistic 2-tuple set is now given (see Algorithm \ref{algo_sa}).
If we consider the graphical example of Figure \ref{fig:simplif}, the linguistic 2-tuple set we obtain is the following (ordered by level):\\
$\{(s_{1}^{3},0),(s_{1}^{5},0),(s_{3}^{5},0),(s_{5}^{9},0),(s_{7}^{9},0),(s_{9}^{17},0),(s_{11}^{17},0)\}$, where $a \gets (s_{1}^{3},0)$, $b \gets (s_{1}^{5},0)$,
$c \gets (s_{3}^{5},0)$, $d \gets (s_{5}^{9},0)$, $e \gets (s_{7}^{9},0)$, $f \gets (s_{9}^{17},0)$ and $g  \gets (s_{11}^{17},0)$.
The last graph of the figure shows the semantics obtained, using the representation algorithm described in~\cite{Herrera08afuzzy}.

\begin{algorithm}[t]
\caption{Simplification algorithm}
\label{algo_sa}
\begin{algorithmic}[1]
\REQUIRE $o$ is a node, $T$ is a binary tree, $o^{\prime}$ is the root node of $T$
\STATE $o^{\prime} \gets (s_0^3,0)$

\FOR {each node $o \in T, o\neq o^{\prime}$}
   \STATE let $(s_i^j,k)$ be the parent node of $o$
   \IF {$o$ is a left child} 
        \STATE $o \gets (s_{2i-1}^{2j-1},0)$
   \ELSE
        \STATE $o \gets (s_{2i+1}^{2j-1},0)$
   \ENDIF
\ENDFOR
\RETURN the set of linguistic 2-tuples, one per node
\end{algorithmic}
\end{algorithm}

\vspace{.3cm}

In a way, this algorithm permits to flatten a binary tree into a 2-tuple set which can be useful to express distances between nodes.
The opposite is also true: a linguistic term set can be expressed through a binary tree.
One of the advantages to perform this flattening is to consider a new dimension in the data of a given problem. This new dimension is the distance
between the possible outcomes (the nodes that can be decisions, choices, preferences, etc.) of the problem and this would allow for a ranking of the outcomes, as if we had a B-tree.
The fact that the level of the linguistic hierarchy is not the same, depending on the node depth, is interesting since it gives a different
granularity level, and, as with Zadeh's granules, it permits to connect a position in the tree and a precision level.

\section{Concluding remarks}

In this paper, we have formally introduced and discussed an approach to deal with unbalanced linguistic term sets.
Our approach is inspired by the 2-tuple fuzzy linguistic representation model from
Herrera and Mart\'inez, but we fully take advantage of the symbolic translations $\alpha$ that become a very important
element to generate the data set.

The 2-tuples of our linguistic model are twofold. Indeed, except the first one and the last one of the partition that have a shape of right-angled triangles,
they all are composed of two \emph{half} 2-tuples: an upside and a downside 2-tuple. The upside
and downside of the 2-tuple are not necessary expressed in the same hierarchy nor level.
Regarding the partitioning phase, there is no need to have all the symbolic translations
equal to zero. This permits to express the non-uniformity of the data much better.

Despite the changes we made, the minimal cover property is fulfilled and proved.
Moreover, the aggregation operators that we redefine give consistent and satisfactory results.
Next steps in future work will be to study other operators, such as comparison, negation, aggregation, implication, etc.

\section*{ACKNOWLEDGEMENT}
\small
This work is partially funded by the French National Research Agency (ANR) under grant number ANR-09-SEGI-012.

\makesubmdate

\bibliographystyle{plain}
\nocite*{}
\bibliography{AbchirTruck}


\makecontacts

\end{document}